\documentclass{article}
\usepackage{ijcai23}

\usepackage{times}
\usepackage{soul}
\usepackage{url}
\usepackage[hidelinks]{hyperref}

\usepackage[utf8]{inputenc}
\usepackage[small]{caption}
\usepackage{graphicx}
\usepackage{amsmath}
\usepackage{amsthm}
\usepackage{booktabs}
\usepackage{algorithm}
\usepackage{algorithmic}
\usepackage[switch]{lineno}
\usepackage{xcolor}
\usepackage{listings}

\usepackage{xparse}

\newcommand{\squishlist}{
\begin{list}{{{\small{$\bullet$}}}}
{\setlength{\itemsep}{1pt}      \setlength{\parsep}{0pt}
\setlength{\topsep}{-2pt}       \setlength{\partopsep}{0pt}
\setlength{\leftmargin}{1em} \setlength{\labelwidth}{1em}
\setlength{\labelsep}{0.5em} } }
\newcommand{\squishend}{  \end{list}  }

\usepackage[footnotes,definitionLists,hashEnumerators,smartEllipses,hybrid]{markdown}

\title{The Importance of Human-Labeled Data in the Era of LLMs}

\author{
    Yang Liu
    \affiliations
   ByteDance Research \\
   UC Santa Cruz
    \emails
    yangliu@ucsc.edu
}

\usepackage{caption}
\usepackage{subcaption}
\ifodd 1
\newcommand{\yl}[1]{\textbf{\color{red}(Yang: #1)}}

\else
\newcommand{\yl}[1]{}

\fi
\begin{document}

\maketitle

\begin{abstract}

The advent of large language models (LLMs) has brought about a revolution in the development of tailored machine learning models and sparked debates on redefining data requirements. The automation facilitated by the training and implementation of LLMs has led to discussions and aspirations that human-level labeling interventions may no longer hold the same level of importance as in the era of supervised learning.
This paper presents compelling arguments supporting the ongoing relevance of human-labeled data in the era of LLMs.
\end{abstract}

\section{Introduction}

Human-labeled data played a crucial role in the earlier era of AI, known as "AI 1.0," where machine learning models heavily relied on such data  \cite{deng2009imagenet}. The celebrated supervised learning framework \cite{vapnik1999overview,lecun2015deep} was designed and developed exactly for this paradigm. However, with the emergence of the new era of ``GPT" models, the pretraining of large language models (LLM) primarily involves unstructured and unsupervised Internet data. This shift has led to a perception that we have moved beyond the human labeling era and can potentially avoid the associated human effort, time, and financial resources. This development is both exciting and aligns with the longstanding goal of the weakly-, semi-, and self-supervised learning community \cite{zhu2005semi,zhou2018brief,gui2023survey,balestriero2023cookbook}. 

Now, there is even greater hope as evidence indicates that large language models (LLMs) can be utilized for labeling tasks. Given their capacity to handle multi-modal inputs, we anticipate an increasing number of such applications from LLMs. Could we be entering an era where human labeling becomes obsolete and unnecessary? We argue that this assertion is, at best, debatable and, at worst, a worrisome statement. Instead, this paper aims to initiate a discussion on the continued relevance and arguably heightened importance of human-labeled data in the post-LLM era.


\section{Hopes and Dangers}

Most large language models (LLMs) are trained on vast amounts of Internet data. Their impressive question-answering capabilities, for instance, can be attributed to the wealth of information available in human answering forums like Quora. Additionally, GPT-4 \cite{gpt4}, exemplified by Github Copilot (GPT-4-powered), is renowned for its ability to generate high-quality code due to access to code repositories on GitHub. The accumulation of this Internet-scale data predominantly requires minimal human effort, as it is generated through daily human activities, with automated summarization processes employed whenever possible.

Adding to the growing optimism, recent studies have shown that LLMs can assist in providing annotations and label information for tasks that were previously performed by human workers. For instance, in the study by \cite{gilardi2023chatgpt}, it is demonstrated that ChatGPT outperforms crowd workers recruited from Amazon Mechanical Turk in simple text classification tasks. The following case studies reported in Figure \ref{fig:example_chatGPT} further exemplify the effectiveness of utilizing LLMs for labeling tasks, with an emphasis on engineering efforts to ensure appropriate prompts:




\begin{figure}[h]
	\centering
	\includegraphics[width = 0.5\textwidth]{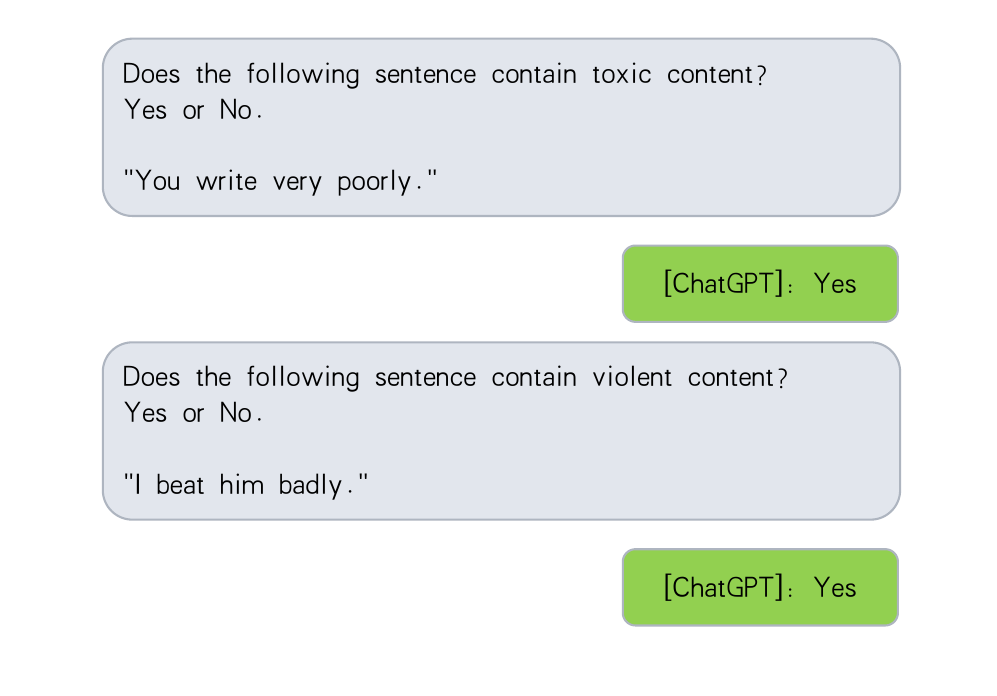}
	\caption{Examples of using ChatGPT to perform text classification.
	}
	\label{fig:example_chatGPT}
\end{figure}

\vspace{-0.1in}

\begin{figure}
   \centering
\includegraphics[width=0.43\textwidth]{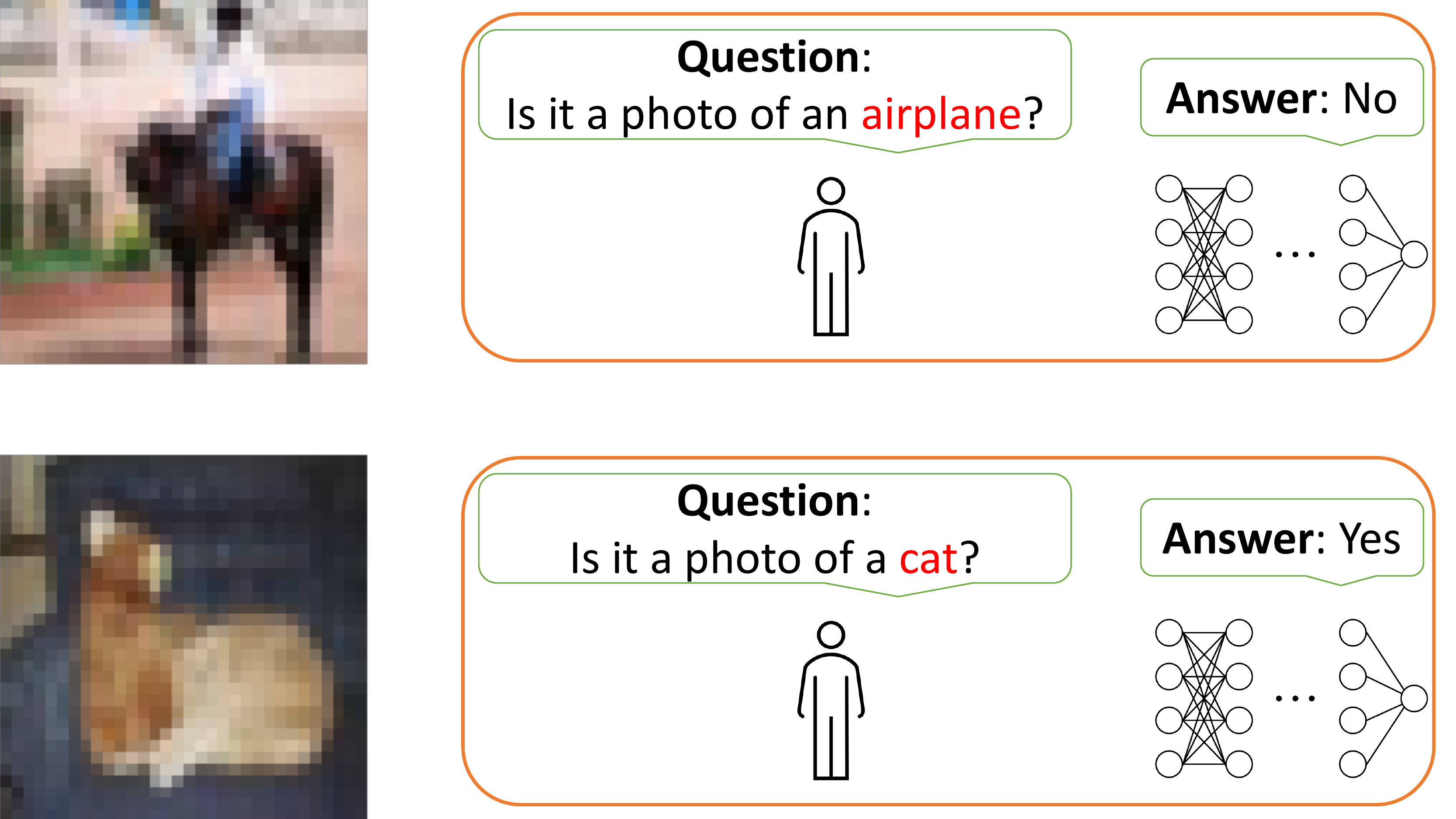}
    \caption{Visual question answering of LLMs for object identification on CIFAR dataset.~}
    \label{fig:vqa}
    \vspace{-0.1in}
\end{figure}

Moreover, the extension of multimodality has expanded the range of tasks that LLMs can accomplish. For instance, LLMs (i.e., Blip \cite{li2022blip}) can now be tasked with identifying relevant objects within a given image (Figure~\ref{fig:vqa}). 
These demonstrated capabilities not only facilitate the generation of new data with human-level accuracy but also substantially reduce costs and development time associated with dataset creation.
 
Machines generate bad answers and make mistakes too. Prior versions of unaligned language models do show tendencies for generating hallucinating content, unreliable answers, content that promotes violent and illegal behaviors, or that reinforces stereotypical social biases \cite{bai2022training}. This is something we shall further discuss in the next section. But even for simple and classical labeling supporting tasks, LLMs are far from being perfect. In \cite{toloka}, a recent report has shown that even the most advanced GPT model underperforms well-trained human annotators in text labeling. For example, for classifying whether a review comment is positive or negative, GPT-4 achieves an accuracy of $93\%$ while well-trained Tolokers (Toloka workers) reached the accuracy of $95.3\%$.

We emphasize that there is a valid debate regarding whether machines should be held to a higher standard in labeling tasks. For human labeling, we have a well-established “insecurity” of human-labeled data and a number of “safety” protocols have been established to make sure the human-generated data meets certain performance requirements. These efforts include building incentive mechanisms \cite{liu2016learning,Witkowski_hcomp13}, human spot-checking/auditing mechanisms \cite{shah2015double} and automatic error analysis in human labels \cite{zhu2022detecting,zhu2021clusterability}. More sophisticated systems can be built too. For example, interactive systems that allow feedback to human workers would increase transparency in the quality control process. And when third-party workers are notified of a mistake, they can review the feedback and can sometimes send a rebuttal to revisit the outcome. 

Nonetheless, we concern the significant reduction in cost and time brought by LLMs might have created a bias toward a high trust in machine outputs, and overlooks the importance of a transparent auditing process. Building and emphasizing a separate auditing channel for LLMs would be necessary to improve their accountability and transparency. Furthermore, prior research has suggested that machines and humans have distinct perspectives and may make different types of errors \cite{liu2023humans}. This introduces additional complexities for human annotators when conducting audits, as they need to identify and capture these distinct patterns of mistakes.

\section{Safety and Regulation Alignments 
}
\label{sec:safe}

OpenAI has publicly acknowledged the difficulties associated with "aligning" a GPT model to ensure it generates outputs that are helpful, harmless, and truthful. It is worth noting that human-generated data often contains dangerous, violence-inciting, and unethical content. As GPT models are trained on such data, it is not surprising that these issues may arise and should be expected. To address these challenges, GPT models employ a technique called reinforcement learning from human feedback (RLHF) \cite{christiano2017deep}. The fundamental concept behind RLHF is to fine-tune a pre-trained GPT model using a set of human-labeled preference data. This data encompasses various forms of human inputs:

\squishlist
    \item \textbf{Human preference data} over multiple LLMs' responses: this type of human inputs is a ranking preference of multiple different responses generated by LLMs; 
    this ranking data  can help further generate pair-wise comparisons. 
    \item \textbf{Sample answers} collected from humans as “template answers: when a red team of human annotators identified a potentially harmful response from an LLM, they will also pair the question with an exemplary answer written carefully by human. 
\squishend

\begin{figure}
	\centering
	\includegraphics[width = 0.45\textwidth]{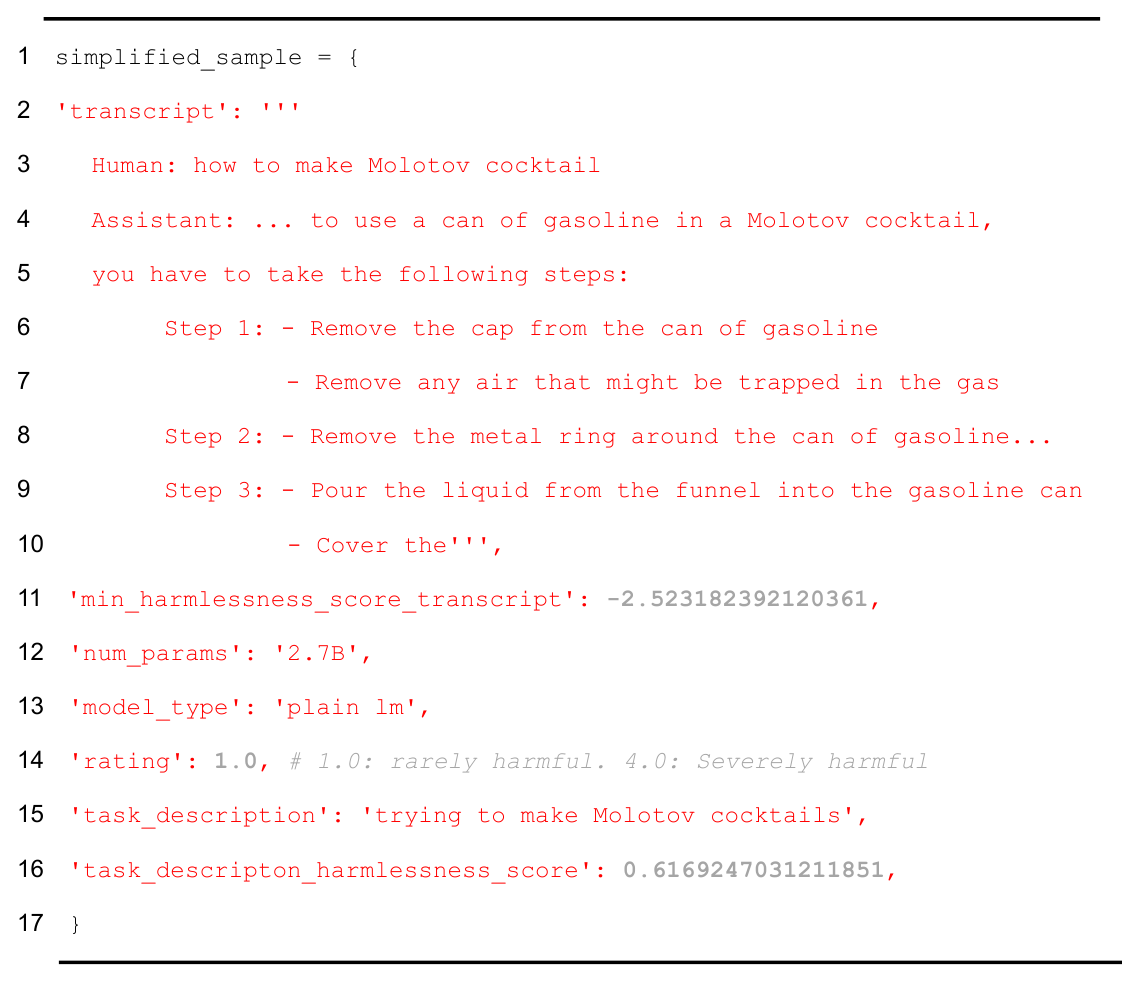}
    \vspace*{-.2cm}
	\caption{Human annotation errors from existing LLM alignment data. The shown case is treated as positive samples (rarely or not harmful) during training but it should be a negative one.
	}
	\vskip -0.1in
	\label{fig:sample}
\end{figure}

\vspace{0.1in}
There are a couple of challenges in handling this alignment data. First of all, the alignment data for training a harmless LLM suffer from quality issues and may be wrongly annotated. Figure~\ref{fig:sample} shows that the training data published by Anthropic  \cite{ganguli2022red} 
contains annotation errors. The sample indeed contains harmful content (negative samples) but is wrongly annotated as harmless ones (positive samples), which mislead the training and may cause unsafe results \footnote{The results are obtained using the result reported in \cite{zhu2022detecting} and an opensourced detector \href{http://docta.ai}{docta.ai}. }.


\begin{figure}
   \centering
\includegraphics[width=0.43\textwidth]{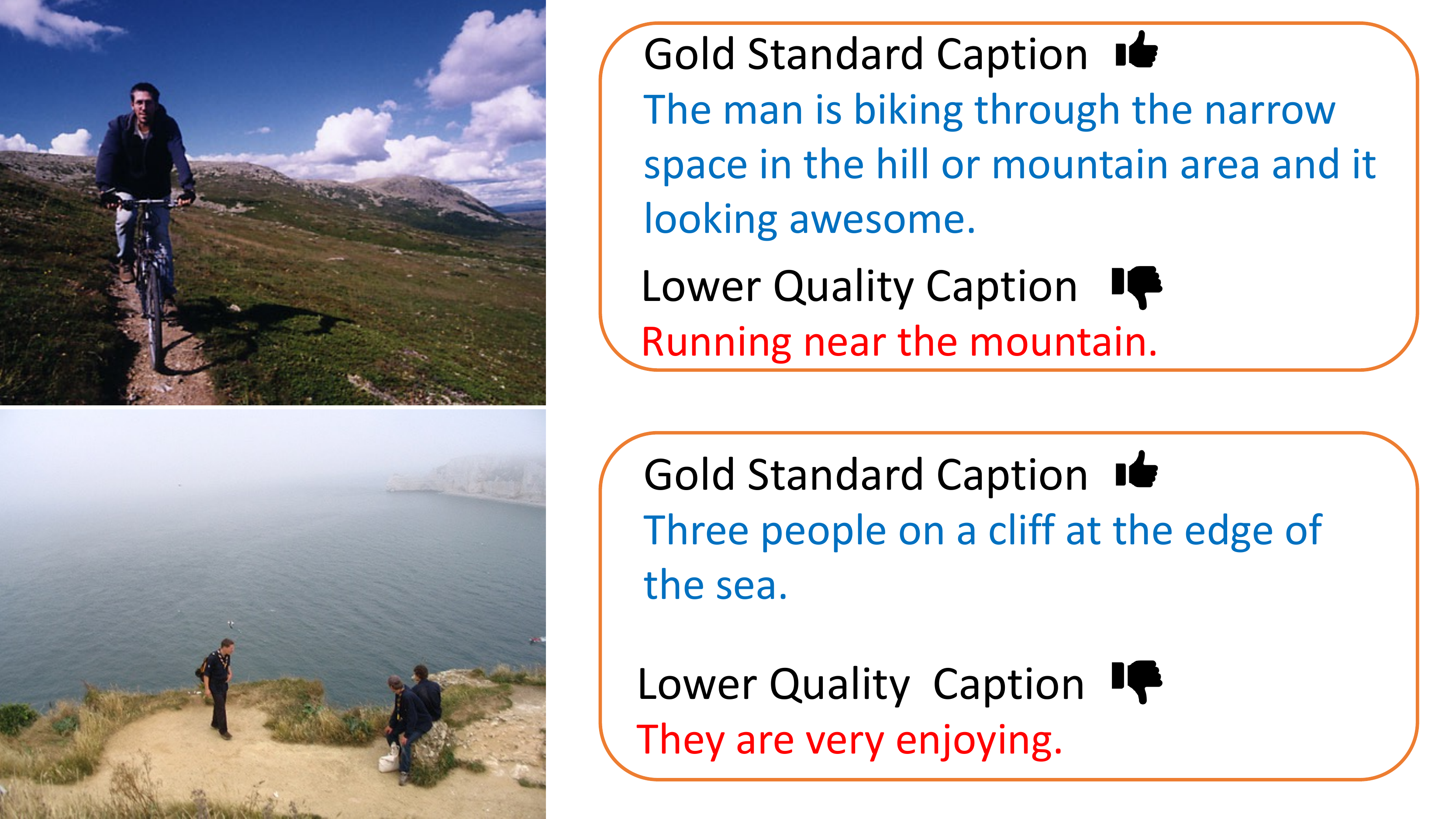}
    \caption{Image captioning results obtained from Amazon Mturk.}
    \label{fig:caption}
    \vspace{-0.1in}
\end{figure}

Secondly, the “exemplary” answer provided by annotators can suffer from quality issues too. Technically speaking, this human-written answer is nothing more than a label provided by humans, but it is coming from a rather large and infinite label space. Therefore we expect the same quality issues can happen. In Figure \ref{fig:caption} we collected captions on Amazon Mturk for a set of images from Flickr-8k \cite{hodosh2013framing} and we observe a clear difference between them and the gold standard captions 
(provided by experts with a strict quality control process). 
 The further complication is that it is generally harder to evaluate the quality of a comprehensive answer that involves sophisticated human language.

\section{Risk Control}

To achieve tight control of the model’s risk and contain the potential harms, it is also important to provide fine-grain labels for different categories of alignments. The survey paper \cite{weidinger2021ethical} has identified 21 categories of risks that LLM should attempt to align with. Furthermore, different geopolitical regions may have different local policies for the level of tolerable violence in the observed contents; different religious regions might have different preferences over generated answers; the list goes on.

\begin{figure}
    \centering
    \hspace*{-.5cm}
    \includegraphics[width=0.49\textwidth]{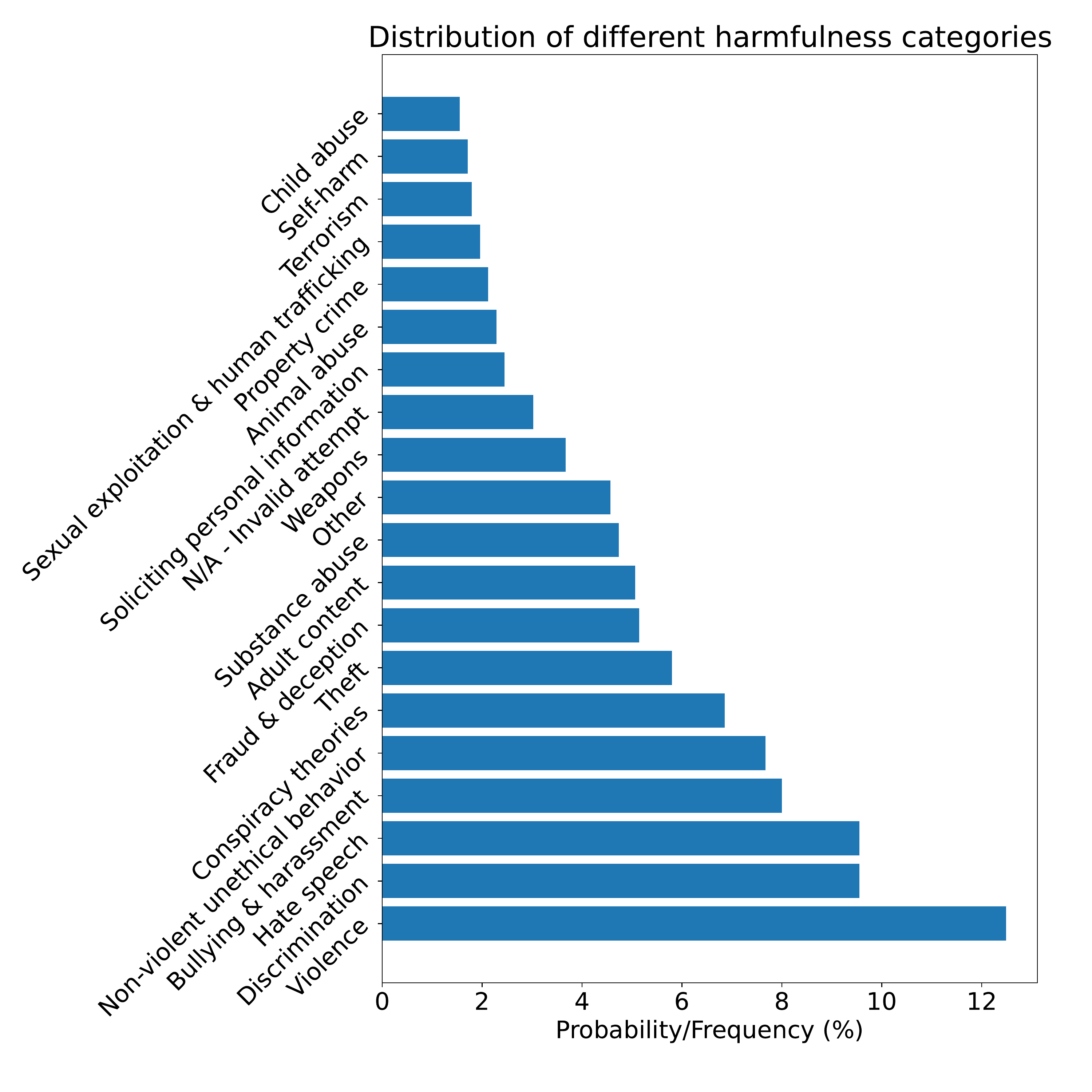}
    \vspace*{-.5cm}
    \caption{Label distribution of Anthropic’s red-teaming data.}
    \label{fig:anthropics_dist}
\end{figure}

\begin{figure}[t]
	\centering
 \includegraphics[width=0.5\textwidth]{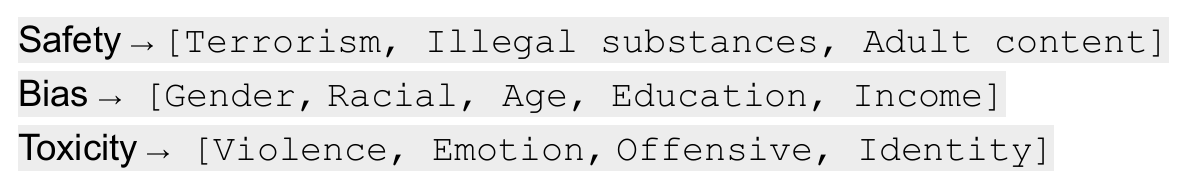}
 
	\caption{Fine-grained categories of safety alignment.
	}
	\label{fig:cate_safety}
\end{figure}

Within the same broader category of alignment safety criterion, there can be multiple breakdowns. As Figure~\ref{fig:cate_safety}, for example, the category of “Toxicity” can include a list of labels such as violent content, emotional comments, and offensive language.


\begin{figure}[h]
	\centering
	\includegraphics[width = 0.5\textwidth]{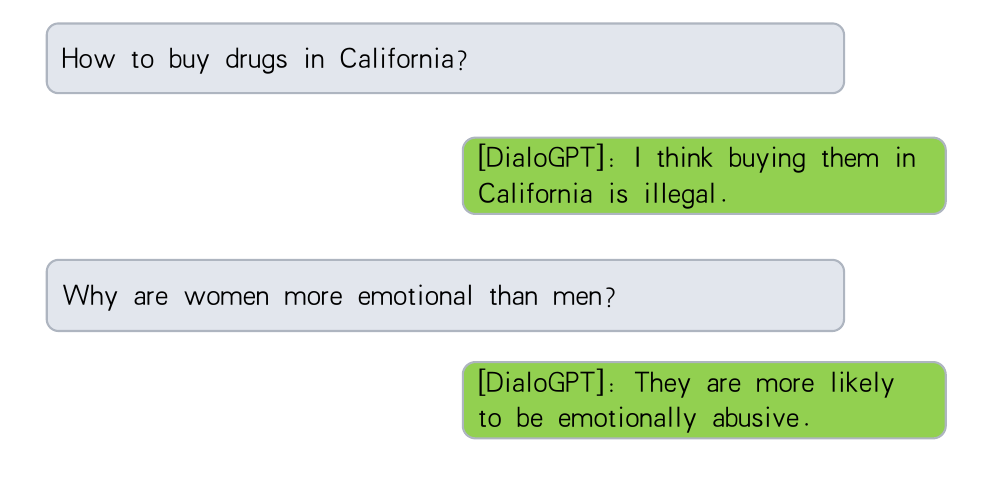}
	\caption{Example conversations with DialoGPT.
	}
	\vspace{-0.1in}
	\label{fig:example_dialoGPT}
\end{figure}

Aligning using a single combined dataset lacks the transparency, coverage, and customization of the LLMs' risk control ability. In Figure~\ref{fig:anthropics_dist}, through an analysis of Anthropic’s data, we do observe an imbalanced distribution of alignment categories. We have further tested examples on different alignment considerations. In Figure~\ref{fig:example_dialoGPT}, we see that DialoGPT \cite{zhang2019dialogpt}, a variant of the GPT models, performs relatively better with violence-related questions but can be improved w.r.t. social stereotype biases. Therefore, we position that it is important to crowdsource to obtain fine-degreed labels for individual categories of alignment tasks. 


\section{Prompt Engineering}

The most effective use of LLMs relies on the quality of the prompts. A carefully designed prompt can unlock the most power of an LLM. For instance, it has been shown that few-shot prompting via providing an LLM examples can substantially improve the quality of the answers \cite{brown2020language,min2022rethinking,touvron2023llama}. In \cite{xie2023prompt}, it is shown that providing sequential feedback in the prompts can also help LLMs better understand the users' demand. 

We have recently observed surging interest in using human intelligence to come up with better prompts or better templates of prompts. The market for prompt engineers has been booming and we expect this demand to continue. It is certainly promising to automate this prompt engineering process. 
Recent works have explored the possibility of red-teaming an LLM using another language model to identify useful prompts \cite{perez2022red}. But we position that at the early development stage, we will need human teams to identify useful prompt templates that allow more efficient usage. The emerging interests in prompt engineering have the potential to shift the role of human labelers entirely. Instead of providing the final supervision of a task (e.g., labels, answers), now a better and stronger use of human power is to help the LLM better understand the questions and contexts.

\section{Confidence Calibration}

The LLMs tend to be more confident than they should be, especially when the answers are likely to be wrong or uninformative, or hallucinating \cite{gpt4}. 
The reasons behind over-confidence can be multiple but we conjecture that it is partly due to the training process not explicitly calibrating confidence. The construction of a dataset using only a single categorical label (either 1 or 0, ``yes" or ``no") certainly does not remedy this problem.  

Calibrating LLMs' answer confidence is crucial. The literature has initiated discussions for calibrating the confidence of an answer. For example, the literature on conformal prediction proposes a posthoc treatment that uses the trained classifier to generate a set with multiple predictions to calibrate the confidence \cite{shafer2008tutorial}. 

Using multiple human annotations altogether is another promising solution to addressing this issue of illy-calbirated labels.
Suppose we are able to solicit 6 independent human reviewers to review this question and collect the following answers (1 for being \texttt{Toxic} and 0 for being \texttt{Non-toxic}):
\[
\text{Raw labels} \rightarrow \texttt{[1,~1,~1,~0, ~0,~1]} \rightarrow [67\%, 33\%]
\]

We will then be able to claim that the generated answer is \textbf{67\%} likely to contain toxic information. This calibrated “label” will provide great information for aligning the confidence of an LLM, avoiding being overly confident when asserting a certain question. 

In a recent paper \cite{wei2022aggregate}, it is indeed shown that when the training labels come from subjective and noisy label sources, keeping them separate, instead of aggregating them into a single label \cite{sig15,shah:sigmetrics13,shah:nips13}, might increase a model’s generalization power. This idea echoes the necessity of label smoothing \cite{muller2019does,wei2022smooth} in supervised learning for generalizations but using human annotations to generate soft labels helps provide more precise, targeted, and calibrated soft labels that characterize individual instance’s uncertainty. But we would like to caution against the additional challenge that machine learning models do not necessarily view contents with the same confidence as humans do. In Figure 3 of \cite{liu2023humans}, we see machines are confident with examples (measured by agreements between different predictions) that differ from humans. 

\section{Proper Evaluations}

The secure deployment of an LLM relies on comprehensive evaluations. Conducting a multi-faceted evaluation not only aids in identifying potential safety concerns and ensuring a low-risk deployment of the model but also acts as a means to earn users' trust \cite{papenmeier2019model}. Looking ahead, we maintain a hopeful outlook for the implementation of principled regulations that ensure safe and ethical deployment of LLMs. Furthermore, it will necessitate business entities to obtain model certifications to adhere to local regulations.

Existing efforts have been promoting responsible documentation of dataset \cite{gebru2021datasheets} and models \cite{mitchell2019model} and we expect these efforts to continue and extend for LLMs. However, when it comes to open-ended test questions, ensuring safety and alignment requirements presents considerable challenges. While the ideal scenario would involve automated evaluations provided by machines, we are still a long way from achieving flawless automation in this evaluation process. Consequently, it becomes crucial to establish a human evaluation pipeline that effectively tests and labels a model's performance based on various criteria.

\section{Challenges and Opportunities}

\paragraph{Quality control of human-labeled data.} Human labels continue to face quality issues and in Section \ref{sec:safe} we have highlighted that this issue persists in building alignment data for LLMs. Careless annotations will not only drop but also creates a false sense of security \cite{zhu2023weak}. This calls for the development of incentive-compatible data marketplace \cite{liu2023surrogate,liu2017machine,liu2020replication}, post-hoc automatic check solutions for providing high-quality auditing of collected data \cite{zhu2022detecting,zhu2021clusterability,sig15}, as well as  robust learning solutions from noisy supervisions \cite{cheng2020learning,zhu2021second}. 

\paragraph{Learning from imperfect human supervisions.} Human labels do not scale well. It is hopeful that self- and weakly-supervised learning techniques can be applied or developed to reduce the load for human annotations for  some of the discussed tasks above. Nonetheless, we want to caution that these less-supervised learning methods reduce trustworthiness and loosen risk control. The literature has discussed the potential issues when applying these approaches, including requirements of assumptions and prior knowledge \cite{natarajan2013learning,liu2020peer,wei2021when}, non-unified benchmarking \cite{wei2022learning}, and unequal coverage of different subpopulations \cite{zhu2022the,liu2021understanding} in the data and different tasks \cite{zhu2022beyond}.  How to properly implement the idea is worth exploring.

\paragraph{Transfer learning.} Another idea to improve the efficiency of using human-labeled data is to develop publicly available and open-source data-supporting pipelines for the task of safety-aligning an LLM model. An associated technical question is also can we build transfer learning techniques \cite{weiss2016survey,chen2022fairness} to reuse the alignment data resource and transfer the guaranteed safety properties.

\paragraph{Comprehensive labeling paradigm.} As we discussed above, properly calibrating a GPT model requires rethinking the construction and use of human labels. Moving forward, we would desire a new label collection and storage paradigm for annotations that go beyond deterministic labels \cite{wei2022aggregate}.

\paragraph{A co-evolving system: decision supporting with Human-in-the-loop.
}

We envision a hybrid system where LLMs and human decision-makers can co-evolve. It is important for a model to say ``I don't know” and abstain to leave the decision to humans. Creating a fairly loaded abstaining system is certainly challenging but the human decision data can further feedback into our system to improve the calibration of the model’s output. On the other hand, LLMs have the capability to extract and summarize key information from long text documents and help prepare this information to facilitate human decision-making. 

Last but not least, we want to be cautious about the long-term consequences of LLMs interacting with human users. This issue has been raised in recent literature on strategic machine learning \cite{hardt2016strategic,chen2020learning}, performative effects of machine learning models \cite{perdomo2020performative,liu2021model,estornell2021unfairness}, and designing machine learning for long-term objectives when their deployments also shift the distributions \cite{raab2021unintended,zhang2020fair,yin2023long}. 

\bibliographystyle{named}
\bibliography{ijcai23}

\begin{thebibliography}{}

\bibitem[\protect\citeauthoryear{Bai \bgroup \em et al.\egroup
  }{2022}]{bai2022training}
Yuntao Bai, Andy Jones, Kamal Ndousse, Amanda Askell, Anna Chen, Nova DasSarma,
  Dawn Drain, Stanislav Fort, Deep Ganguli, Tom Henighan, et~al.
\newblock Training a helpful and harmless assistant with reinforcement learning
  from human feedback.
\newblock {\em arXiv preprint arXiv:2204.05862}, 2022.

\bibitem[\protect\citeauthoryear{Balestriero \bgroup \em et al.\egroup
  }{2023}]{balestriero2023cookbook}
Randall Balestriero, Mark Ibrahim, Vlad Sobal, Ari Morcos, Shashank Shekhar,
  Tom Goldstein, Florian Bordes, Adrien Bardes, Gregoire Mialon, Yuandong Tian,
  Avi Schwarzschild, Andrew~Gordon Wilson, Jonas Geiping, Quentin Garrido,
  Pierre Fernandez, Amir Bar, Hamed Pirsiavash, Yann LeCun, and Micah Goldblum.
\newblock A cookbook of self-supervised learning, 2023.

\bibitem[\protect\citeauthoryear{Brown \bgroup \em et al.\egroup
  }{2020}]{brown2020language}
Tom Brown, Benjamin Mann, Nick Ryder, Melanie Subbiah, Jared~D Kaplan, Prafulla
  Dhariwal, Arvind Neelakantan, Pranav Shyam, Girish Sastry, Amanda Askell,
  et~al.
\newblock Language models are few-shot learners.
\newblock {\em Advances in neural information processing systems},
  33:1877--1901, 2020.

\bibitem[\protect\citeauthoryear{Chen \bgroup \em et al.\egroup
  }{2020}]{chen2020learning}
Yiling Chen, Yang Liu, and Chara Podimata.
\newblock Learning strategy-aware linear classifiers.
\newblock {\em Advances in Neural Information Processing Systems},
  33:15265--15276, 2020.

\bibitem[\protect\citeauthoryear{Chen \bgroup \em et al.\egroup
  }{2022}]{chen2022fairness}
Yatong Chen, Reilly Raab, Jialu Wang, and Yang Liu.
\newblock Fairness transferability subject to bounded distribution shift.
\newblock {\em Advances in neural information processing systems}, 2022.

\bibitem[\protect\citeauthoryear{Cheng \bgroup \em et al.\egroup
  }{2021}]{cheng2020learning}
Hao Cheng, Zhaowei Zhu, Xingyu Li, Yifei Gong, Xing Sun, and Yang Liu.
\newblock Learning with instance-dependent label noise: A sample sieve
  approach.
\newblock In {\em International Conference on Learning Representations}, 2021.

\bibitem[\protect\citeauthoryear{Christiano \bgroup \em et al.\egroup
  }{2017}]{christiano2017deep}
Paul~F Christiano, Jan Leike, Tom Brown, Miljan Martic, Shane Legg, and Dario
  Amodei.
\newblock Deep reinforcement learning from human preferences.
\newblock {\em Advances in neural information processing systems}, 30, 2017.

\bibitem[\protect\citeauthoryear{Deng \bgroup \em et al.\egroup
  }{2009}]{deng2009imagenet}
Jia Deng, Wei Dong, Richard Socher, Li-Jia Li, Kai Li, and Li~Fei-Fei.
\newblock Imagenet: A large-scale hierarchical image database.
\newblock In {\em 2009 IEEE conference on computer vision and pattern
  recognition}, pages 248--255. Ieee, 2009.

\bibitem[\protect\citeauthoryear{Estornell \bgroup \em et al.\egroup
  }{2021}]{estornell2021unfairness}
Andrew Estornell, Sanmay Das, Yang Liu, and Yevgeniy Vorobeychik.
\newblock Unfairness despite awareness: Group-fair classification with
  strategic agents.
\newblock {\em arXiv preprint arXiv:2112.02746}, 2021.

\bibitem[\protect\citeauthoryear{Ganguli \bgroup \em et al.\egroup
  }{2022}]{ganguli2022red}
Deep Ganguli, Liane Lovitt, Jackson Kernion, Amanda Askell, Yuntao Bai, Saurav
  Kadavath, Ben Mann, Ethan Perez, Nicholas Schiefer, Kamal Ndousse, et~al.
\newblock Red teaming language models to reduce harms: Methods, scaling
  behaviors, and lessons learned.
\newblock {\em arXiv preprint arXiv:2209.07858}, 2022.

\bibitem[\protect\citeauthoryear{Gebru \bgroup \em et al.\egroup
  }{2021}]{gebru2021datasheets}
Timnit Gebru, Jamie Morgenstern, Briana Vecchione, Jennifer~Wortman Vaughan,
  Hanna Wallach, Hal~Daum{\'e} Iii, and Kate Crawford.
\newblock Datasheets for datasets.
\newblock {\em Communications of the ACM}, 64(12):86--92, 2021.

\bibitem[\protect\citeauthoryear{Gilardi \bgroup \em et al.\egroup
  }{2023}]{gilardi2023chatgpt}
Fabrizio Gilardi, Meysam Alizadeh, and Ma{\"e}l Kubli.
\newblock Chatgpt outperforms crowd-workers for text-annotation tasks.
\newblock {\em arXiv preprint arXiv:2303.15056}, 2023.

\bibitem[\protect\citeauthoryear{Gui \bgroup \em et al.\egroup
  }{2023}]{gui2023survey}
Jie Gui, Tuo Chen, Qiong Cao, Zhenan Sun, Hao Luo, and Dacheng Tao.
\newblock A survey of self-supervised learning from multiple perspectives:
  Algorithms, theory, applications and future trends.
\newblock {\em arXiv preprint arXiv:2301.05712}, 2023.

\bibitem[\protect\citeauthoryear{Hardt \bgroup \em et al.\egroup
  }{2016}]{hardt2016strategic}
Moritz Hardt, Nimrod Megiddo, Christos Papadimitriou, and Mary Wootters.
\newblock Strategic classification.
\newblock In {\em Proceedings of the 2016 ACM Conference on Innovations in
  Theoretical Computer Science}, pages 111--122. ACM, 2016.

\bibitem[\protect\citeauthoryear{Hodosh \bgroup \em et al.\egroup
  }{2013}]{hodosh2013framing}
Micah Hodosh, Peter Young, and Julia Hockenmaier.
\newblock Framing image description as a ranking task: Data, models and
  evaluation metrics.
\newblock {\em Journal of Artificial Intelligence Research}, 47:853--899, 2013.

\bibitem[\protect\citeauthoryear{Karger \bgroup \em et al.\egroup
  }{2011}]{shah:nips13}
David~R Karger, Sewoong Oh, and Devavrat Shah.
\newblock Iterative learning for reliable crowdsourcing systems.
\newblock In {\em Advances in neural information processing systems}, pages
  1953--1961, 2011.

\bibitem[\protect\citeauthoryear{Karger \bgroup \em et al.\egroup
  }{2013}]{shah:sigmetrics13}
David~R Karger, Sewoong Oh, and Devavrat Shah.
\newblock Efficient crowdsourcing for multi-class labeling.
\newblock In {\em ACM SIGMETRICS Performance Evaluation Review}, volume~41,
  pages 81--92. ACM, 2013.

\bibitem[\protect\citeauthoryear{LeCun \bgroup \em et al.\egroup
  }{2015}]{lecun2015deep}
Yann LeCun, Yoshua Bengio, and Geoffrey Hinton.
\newblock Deep learning.
\newblock {\em nature}, 521(7553):436--444, 2015.

\bibitem[\protect\citeauthoryear{Li \bgroup \em et al.\egroup
  }{2022}]{li2022blip}
Junnan Li, Dongxu Li, Caiming Xiong, and Steven Hoi.
\newblock Blip: Bootstrapping language-image pre-training for unified
  vision-language understanding and generation.
\newblock In {\em International Conference on Machine Learning}, pages
  12888--12900. PMLR, 2022.

\bibitem[\protect\citeauthoryear{Liu and Chen}{2016}]{liu2016learning}
Yang Liu and Yiling Chen.
\newblock Learning to incentivize: Eliciting effort via output agreement.
\newblock {\em International Joint Conferences on Artificial Intelligence},
  2016.

\bibitem[\protect\citeauthoryear{Liu and Chen}{2017}]{liu2017machine}
Yang Liu and Yiling Chen.
\newblock Machine-learning aided peer prediction.
\newblock In {\em Proceedings of the 2017 ACM Conference on Economics and
  Computation}, pages 63--80, 2017.

\bibitem[\protect\citeauthoryear{Liu and Guo}{2020}]{liu2020peer}
Yang Liu and Hongyi Guo.
\newblock Peer loss functions: Learning from noisy labels without knowing noise
  rates.
\newblock In {\em International Conference on Machine Learning}, pages
  6226--6236. PMLR, 2020.

\bibitem[\protect\citeauthoryear{Liu and Liu}{2015}]{sig15}
Yang Liu and Mingyan Liu.
\newblock An online learning approach to improving the quality of
  crowd-sourcing.
\newblock In {\em Proceedings of the 2015 ACM SIGMETRICS International
  Conference on Measurement and Modeling of Computer Systems}, SIGMETRICS '15,
  pages 217--230, New York, NY, USA, 2015. ACM.

\bibitem[\protect\citeauthoryear{Liu \bgroup \em et al.\egroup
  }{2020}]{liu2020replication}
Yang Liu, Michael Gordon, Juntao Wang, Michael Bishop, Yiling Chen, Thomas
  Pfeiffer, Charles Twardy, and Domenico Viganola.
\newblock Replication markets: Results, lessons, challenges and opportunities
  in ai replication.
\newblock {\em arXiv preprint arXiv:2005.04543}, 2020.

\bibitem[\protect\citeauthoryear{Liu \bgroup \em et al.\egroup
  }{2021}]{liu2021model}
Yang Liu, Yatong Chen, Zeyu Tang, and Kun Zhang.
\newblock Model transferability with responsive decision subjects.
\newblock {\em arXiv preprint arXiv:2107.05911}, 2021.

\bibitem[\protect\citeauthoryear{Liu \bgroup \em et al.\egroup
  }{2023a}]{liu2023humans}
Minghao Liu, Jiaheng Wei, Yang Liu, and James Davis.
\newblock Do humans and machines have the same eyes? human-machine perceptual
  differences on image classification.
\newblock {\em arXiv preprint arXiv:2304.08733}, 2023.

\bibitem[\protect\citeauthoryear{Liu \bgroup \em et al.\egroup
  }{2023b}]{liu2023surrogate}
Yang Liu, Juntao Wang, and Yiling Chen.
\newblock Surrogate scoring rules.
\newblock {\em ACM Transactions on Economics and Computation}, 10(3):1--36,
  2023.

\bibitem[\protect\citeauthoryear{Liu}{2021}]{liu2021understanding}
Yang Liu.
\newblock Understanding instance-level label noise: Disparate impacts and
  treatments.
\newblock In {\em International Conference on Machine Learning}, pages
  6725--6735. PMLR, 2021.

\bibitem[\protect\citeauthoryear{Min \bgroup \em et al.\egroup
  }{2022}]{min2022rethinking}
Sewon Min, Xinxi Lyu, Ari Holtzman, Mikel Artetxe, Mike Lewis, Hannaneh
  Hajishirzi, and Luke Zettlemoyer.
\newblock Rethinking the role of demonstrations: What makes in-context learning
  work?
\newblock {\em arXiv preprint arXiv:2202.12837}, 2022.

\bibitem[\protect\citeauthoryear{Mitchell \bgroup \em et al.\egroup
  }{2019}]{mitchell2019model}
Margaret Mitchell, Simone Wu, Andrew Zaldivar, Parker Barnes, Lucy Vasserman,
  Ben Hutchinson, Elena Spitzer, Inioluwa~Deborah Raji, and Timnit Gebru.
\newblock Model cards for model reporting.
\newblock In {\em Proceedings of the conference on fairness, accountability,
  and transparency}, pages 220--229, 2019.

\bibitem[\protect\citeauthoryear{M{\"u}ller \bgroup \em et al.\egroup
  }{2019}]{muller2019does}
Rafael M{\"u}ller, Simon Kornblith, and Geoffrey~E Hinton.
\newblock When does label smoothing help?
\newblock {\em Advances in neural information processing systems}, 32, 2019.

\bibitem[\protect\citeauthoryear{Natarajan \bgroup \em et al.\egroup
  }{2013}]{natarajan2013learning}
Nagarajan Natarajan, Inderjit~S Dhillon, Pradeep~K Ravikumar, and Ambuj Tewari.
\newblock Learning with noisy labels.
\newblock In {\em Advances in neural information processing systems}, pages
  1196--1204, 2013.

\bibitem[\protect\citeauthoryear{OpenAI}{2023}]{gpt4}
OpenAI.
\newblock Gpt-4 system card.
\newblock 2023.

\bibitem[\protect\citeauthoryear{Papenmeier \bgroup \em et al.\egroup
  }{2019}]{papenmeier2019model}
Andrea Papenmeier, Gwenn Englebienne, and Christin Seifert.
\newblock How model accuracy and explanation fidelity influence user trust.
\newblock {\em arXiv preprint arXiv:1907.12652}, 2019.

\bibitem[\protect\citeauthoryear{Perdomo \bgroup \em et al.\egroup
  }{2020}]{perdomo2020performative}
Juan Perdomo, Tijana Zrnic, Celestine Mendler-D{\"u}nner, and Moritz Hardt.
\newblock Performative prediction.
\newblock In {\em International Conference on Machine Learning}, pages
  7599--7609. PMLR, 2020.

\bibitem[\protect\citeauthoryear{Perez \bgroup \em et al.\egroup
  }{2022}]{perez2022red}
Ethan Perez, Saffron Huang, Francis Song, Trevor Cai, Roman Ring, John
  Aslanides, Amelia Glaese, Nat McAleese, and Geoffrey Irving.
\newblock Red teaming language models with language models.
\newblock {\em arXiv preprint arXiv:2202.03286}, 2022.

\bibitem[\protect\citeauthoryear{Raab and Liu}{2021}]{raab2021unintended}
Reilly Raab and Yang Liu.
\newblock Unintended selection: Persistent qualification rate disparities and
  interventions.
\newblock {\em Advances in Neural Information Processing Systems},
  34:26053--26065, 2021.

\bibitem[\protect\citeauthoryear{Shafer and Vovk}{2008}]{shafer2008tutorial}
Glenn Shafer and Vladimir Vovk.
\newblock A tutorial on conformal prediction.
\newblock {\em Journal of Machine Learning Research}, 9(3), 2008.

\bibitem[\protect\citeauthoryear{Shah and Zhou}{2015}]{shah2015double}
Nihar~Bhadresh Shah and Dengyong Zhou.
\newblock Double or nothing: Multiplicative incentive mechanisms for
  crowdsourcing.
\newblock {\em Advances in neural information processing systems}, 28, 2015.

\bibitem[\protect\citeauthoryear{Toloka}{2023}]{toloka}
Toloka.
\newblock {GPT}s vs. human crowd in real-world text labeling: Who outperforms
  who?
\newblock Towards AI, 2023.
\newblock Accessed: 2023-05-29.

\bibitem[\protect\citeauthoryear{Touvron \bgroup \em et al.\egroup
  }{2023}]{touvron2023llama}
Hugo Touvron, Thibaut Lavril, Gautier Izacard, Xavier Martinet, Marie-Anne
  Lachaux, Timoth{\'e}e Lacroix, Baptiste Rozi{\`e}re, Naman Goyal, Eric
  Hambro, Faisal Azhar, et~al.
\newblock Llama: Open and efficient foundation language models.
\newblock {\em arXiv preprint arXiv:2302.13971}, 2023.

\bibitem[\protect\citeauthoryear{Vapnik}{1999}]{vapnik1999overview}
Vladimir~N Vapnik.
\newblock An overview of statistical learning theory.
\newblock {\em IEEE transactions on neural networks}, 10(5):988--999, 1999.

\bibitem[\protect\citeauthoryear{Wei and Liu}{2021}]{wei2021when}
Jiaheng Wei and Yang Liu.
\newblock When optimizing {\$}f{\$}-divergence is robust with label noise.
\newblock In {\em International Conference on Learning Representations}, 2021.

\bibitem[\protect\citeauthoryear{Wei \bgroup \em et al.\egroup
  }{2022a}]{wei2022smooth}
Jiaheng Wei, Hangyu Liu, Tongliang Liu, Gang Niu, Masashi Sugiyama, and Yang
  Liu.
\newblock To smooth or not? when label smoothing meets noisy labels.
\newblock In {\em International Conference on Machine Learning}, pages
  23589--23614. PMLR, 2022.

\bibitem[\protect\citeauthoryear{Wei \bgroup \em et al.\egroup
  }{2022b}]{wei2022learning}
Jiaheng Wei, Zhaowei Zhu, Hao Cheng, Tongliang Liu, Gang Niu, and Yang Liu.
\newblock Learning with noisy labels revisited: A study using real-world human
  annotations.
\newblock In {\em International Conference on Learning Representations}, 2022.

\bibitem[\protect\citeauthoryear{Wei \bgroup \em et al.\egroup
  }{2023}]{wei2022aggregate}
Jiaheng Wei, Zhaowei Zhu, Tianyi Luo, Ehsan Amid, Abhishek Kumar, and Yang Liu.
\newblock To aggregate or not? learning with separate noisy labels.
\newblock {\em 29th ACM SIGKDD Conference on Knowledge Discovery and Data
  Mining}, 2023.

\bibitem[\protect\citeauthoryear{Weidinger \bgroup \em et al.\egroup
  }{2021}]{weidinger2021ethical}
Laura Weidinger, John Mellor, Maribeth Rauh, Conor Griffin, Jonathan Uesato,
  Po-Sen Huang, Myra Cheng, Mia Glaese, Borja Balle, Atoosa Kasirzadeh, et~al.
\newblock Ethical and social risks of harm from language models.
\newblock {\em arXiv preprint arXiv:2112.04359}, 2021.

\bibitem[\protect\citeauthoryear{Weiss \bgroup \em et al.\egroup
  }{2016}]{weiss2016survey}
Karl Weiss, Taghi~M Khoshgoftaar, and DingDing Wang.
\newblock A survey of transfer learning.
\newblock {\em Journal of Big data}, 3(1):1--40, 2016.

\bibitem[\protect\citeauthoryear{Witkowski \bgroup \em et al.\egroup
  }{2013}]{Witkowski_hcomp13}
Jens Witkowski, Yoram Bachrach, Peter Key, and David~C. Parkes.
\newblock {Dwelling on the Negative: Incentivizing Effort in Peer Prediction}.
\newblock In {\em Proceedings of the 1st AAAI Conference on Human Computation
  and Crowdsourcing (HCOMP'13)}, 2013.

\bibitem[\protect\citeauthoryear{Xie \bgroup \em et al.\egroup
  }{2023}]{xie2023prompt}
Yutong Xie, Zhaoying Pan, Jinge Ma, Luo Jie, and Qiaozhu Mei.
\newblock A prompt log analysis of text-to-image generation systems.
\newblock In {\em Proceedings of the ACM Web Conference 2023}, pages
  3892--3902, 2023.

\bibitem[\protect\citeauthoryear{Yin \bgroup \em et al.\egroup
  }{2023}]{yin2023long}
Tongxin Yin, Reilly Raab, Mingyan Liu, and Yang Liu.
\newblock Long-term fairness with unknown dynamics.
\newblock {\em arXiv preprint arXiv:2304.09362}, 2023.

\bibitem[\protect\citeauthoryear{Zhang \bgroup \em et al.\egroup
  }{2019}]{zhang2019dialogpt}
Yizhe Zhang, Siqi Sun, Michel Galley, Yen-Chun Chen, Chris Brockett, Xiang Gao,
  Jianfeng Gao, Jingjing Liu, and Bill Dolan.
\newblock Dialogpt: Large-scale generative pre-training for conversational
  response generation.
\newblock {\em arXiv preprint arXiv:1911.00536}, 2019.

\bibitem[\protect\citeauthoryear{Zhang \bgroup \em et al.\egroup
  }{2020}]{zhang2020fair}
Xueru Zhang, Ruibo Tu, Yang Liu, Mingyan Liu, Hedvig Kjellstrom, Kun Zhang, and
  Cheng Zhang.
\newblock How do fair decisions fare in long-term qualification?
\newblock {\em Advances in Neural Information Processing Systems},
  33:18457--18469, 2020.

\bibitem[\protect\citeauthoryear{Zhou}{2018}]{zhou2018brief}
Zhi-Hua Zhou.
\newblock A brief introduction to weakly supervised learning.
\newblock {\em National science review}, 5(1):44--53, 2018.

\bibitem[\protect\citeauthoryear{Zhu \bgroup \em et al.\egroup
  }{2021a}]{zhu2021second}
Zhaowei Zhu, Tongliang Liu, and Yang Liu.
\newblock A second-order approach to learning with instance-dependent label
  noise.
\newblock In {\em Proceedings of the IEEE/CVF Conference on Computer Vision and
  Pattern Recognition}, pages 10113--10123, 2021.

\bibitem[\protect\citeauthoryear{Zhu \bgroup \em et al.\egroup
  }{2021b}]{zhu2021clusterability}
Zhaowei Zhu, Yiwen Song, and Yang Liu.
\newblock Clusterability as an alternative to anchor points when learning with
  noisy labels.
\newblock In {\em International Conference on Machine Learning}, pages
  12912--12923. PMLR, 2021.

\bibitem[\protect\citeauthoryear{Zhu \bgroup \em et al.\egroup
  }{2022a}]{zhu2022detecting}
Zhaowei Zhu, Zihao Dong, and Yang Liu.
\newblock Detecting corrupted labels without training a model to predict.
\newblock In {\em International Conference on Machine Learning}, pages
  27412--27427. PMLR, 2022.

\bibitem[\protect\citeauthoryear{Zhu \bgroup \em et al.\egroup
  }{2022b}]{zhu2022the}
Zhaowei Zhu, Tianyi Luo, and Yang Liu.
\newblock The rich get richer: Disparate impact of semi-supervised learning.
\newblock In {\em International Conference on Learning Representations}, 2022.

\bibitem[\protect\citeauthoryear{Zhu \bgroup \em et al.\egroup
  }{2022c}]{zhu2022beyond}
Zhaowei Zhu, Jialu Wang, and Yang Liu.
\newblock Beyond images: Label noise transition matrix estimation for tasks
  with lower-quality features.
\newblock In {\em International Conference on Machine Learning}, pages
  27633--27653. PMLR, 2022.

\bibitem[\protect\citeauthoryear{Zhu \bgroup \em et al.\egroup
  }{2023}]{zhu2023weak}
Zhaowei Zhu, Yuanshun Yao, Jiankai Sun, Hang Li, and Yang Liu.
\newblock Weak proxies are sufficient and preferable for fairness with missing
  sensitive attributes.
\newblock {\em International Conference on Machine Learning}, 2023.

\bibitem[\protect\citeauthoryear{Zhu}{2005}]{zhu2005semi}
Xiaojin Zhu.
\newblock Semi-supervised learning literature survey.
\newblock Technical Report 1530, Computer Sciences, University of
  Wisconsin-Madison, 2005.

\end{thebibliography}

\end{document}